\documentclass[letterpaper]{article} 
\usepackage{aaai2026}  
\usepackage{times}  
\usepackage{helvet}  
\usepackage{courier}  
\usepackage[hyphens]{url}  
\usepackage{graphicx} 
\usepackage{natbib}  
\usepackage{caption} 
\usepackage{algorithm}
\usepackage{algorithmic}
\usepackage{multirow}
\usepackage{amsmath}
\usepackage{booktabs}
\usepackage{amsfonts}
\usepackage{subcaption}
\usepackage{xcolor}

\usepackage{caption}
\usepackage{graphicx}
\usepackage{mathrsfs}
\usepackage{pifont}
\usepackage{times}
\usepackage{helvet}
\usepackage{courier}
\usepackage{newfloat}
\usepackage{listings}
\DeclareCaptionStyle{ruled}{labelfont=normalfont,labelsep=colon,strut=off} 
\floatstyle{ruled}
\newfloat{listing}{tb}{lst}{}
\floatname{listing}{Listing}
\nocopyright

\title{Temporal-Conditional Referring Video Object Segmentation with Noise-Free Text-to-Video Diffusion Model}
\author{
    Ruixin Zhang,
    Jiaqing Fan\thanks{Corresponding author.},
    Yifan Liao,
    Qian Qiao,
    Fanzhang Li
}
\affiliations{
    School of Computer Science and Technology, Soochow University\\

    \texttt{\small 20245227041@stu.suda.edu.cn,joeqian@aliyun.com} \\
  \texttt{\small\{jqfan,lfzh\}@suda.edu.cn} \\
  Code: \url{https://github.com/qianqiaoai/HCD} 
}

\usepackage{bibentry}

\begin{document}

\maketitle

\begin{abstract}
Referring Video Object Segmentation (RVOS) aims to segment specific objects in a video according to textual descriptions. We observe that recent RVOS approaches often place excessive emphasis on feature extraction and temporal modeling, while relatively neglecting the design of the segmentation head. In fact, there remains considerable room for improvement in segmentation head design.
To address this, we propose a Temporal-Conditional Referring Video Object Segmentation model, which innovatively integrates existing segmentation methods to effectively enhance boundary segmentation capability. Furthermore, our model leverages a text-to-video diffusion model for feature extraction. On top of this, we remove the traditional noise prediction module to avoid the randomness of noise from degrading segmentation accuracy, thereby simplifying the model while improving performance.
Finally, to overcome the limited feature extraction capability of the VAE, we design a Temporal Context Mask Refinement (TCMR) module, which significantly improves segmentation quality without introducing complex designs. We evaluate our method on four public RVOS benchmarks, where it consistently achieves state-of-the-art performance.
\end{abstract}
\section{Introduction}
Referring Video Object Segmentation (RVOS) aims to accurately segment specific target objects in videos based on natural language descriptions. As a multimodal task that integrates language understanding and visual perception, RVOS requires models to possess not only strong semantic comprehension of language but also effective cross-modal fusion capabilities between linguistic and visual modalities. In recent years, RVOS has garnered significant attention in the field of computer vision, particularly due to the superior performance of Transformer-based architectures in multimodal representation learning.
\begin{figure}[t]
\centering
\includegraphics[width=0.9\columnwidth]{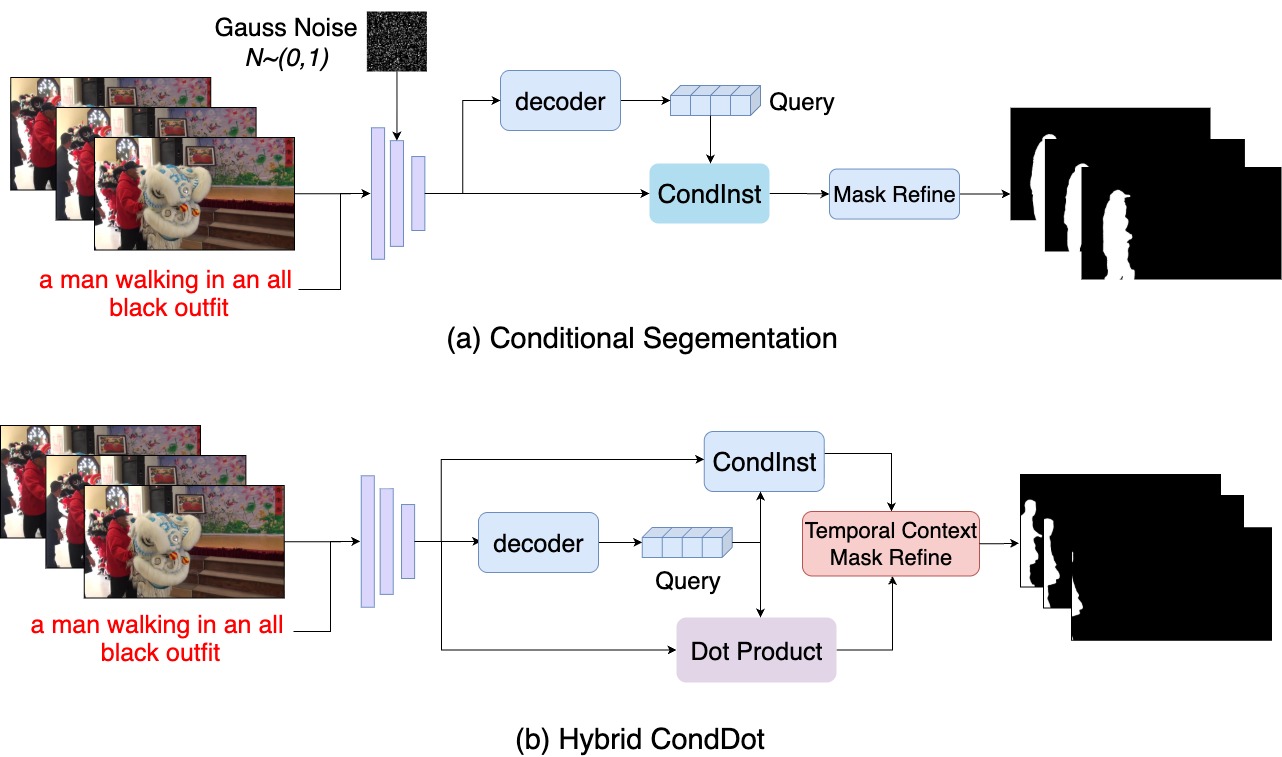} 
\caption{(a) Conditional methods use conditional kernels constructed from the decoded features for target segmentation. (b) Our method integrates existing segmentation methods to concurrently predict instance masks.
  }
\label{fig1}
\end{figure}
Most state-of-the-art RVOS methods~\cite{MANet, ReferDINO} are built upon the DEtection TRansformer (DETR) framework~\cite{DETR}. On one hand, DETR offers a clean and efficient architecture with excellent scalability; on the other hand, its inherent attention mechanism is naturally suited for aligning and interacting between language and vision. Based on this foundation, substantial progress has been made in multimodal feature extraction, semantic interpretation, and cross-modal interaction. For instance, VD-IT~\cite{VD-IT} pioneers the incorporation of diffusion models~\cite{DDPM, stableDiffusion, DIT} into RVOS by leveraging the powerful representation capacity of diffusion models for visual feature extraction, demonstrating impressive temporal modeling capabilities. This approach highlights the great potential of diffusion models in video object segmentation tasks.\\
Despite advancements in video understanding, current methods still face significant challenges in accurately segmenting targets under complex scenarios. Our analysis indicates that the primary bottleneck lies in the segmentation module itself, which has received comparatively little innovation. Existing methods often overemphasize the enhancement of video-language understanding while neglecting the pivotal role that segmentation strategies play in overall performance. In particular, many RVOS models adopt CondInst~\cite{CondInst} as the segmentation head. CondInst generates target masks by dynamically predicting per-layer weights from decoder features and using them to produce object masks. Its advantages lie in its lightweight design, low parameter count, and strong generalization. However, its main drawback is that rich decoder features are compressed into low-dimensional weight vectors, leading to suboptimal utilization of semantic information. In contrast, DETR-based models typically generate masks through dot-product operations between decoder outputs and image features, which allows for better preservation of spatial details, though such approaches struggle to capture complex semantic interactions effectively.\\
To address the limitations of existing approaches, we propose a novel segmentation head, Hybrid CondDot (HCD), incorporating a Dual-path Complementary Enhancement Mechanism. HCD combines CondInst’s dynamic weight adaptation with the semantic richness of dot-product-based mask generation, enabling efficient object shape adaptation and fine-grained spatial semantics, thereby improving segmentation accuracy and robustness.
We further analyze Gaussian noise injection~\cite{referEverything, VD-IT}, commonly used in diffusion models. While beneficial for generative tasks, noise disrupts the spatial structure critical for segmentation, causing unstable training and degraded performance. We therefore remove the noise prediction module, simplifying the architecture and improving stability and output consistency.
Lastly, prior work often employs VAEs~\cite{VAE} for mask refinement, but VAEs struggle to model temporal dependencies. We replace them with a lightweight Temporal Context Mask Refinement Module (TCMR), which explicitly captures inter-frame dependencies to produce temporally coherent and refined masks. Our approach enhances temporal awareness, semantic richness, and robustness under dynamic video conditions with minimal computational overhead.
The contributions of our paper can be summarized as follows:
\begin{itemize}

\item We propose HCD, a novel segmentation method based on a dual-path complementary mechanism that effectively integrates the strengths of existing segmentation techniques. This approach enhances the model’s semantic understanding and global detail capture capabilities, significantly improving segmentation performance.
\item Informed by the detrimental effect of noise, we remove the traditional noise prediction component. This not only streamlines the model architecture but also improves overall performance.
\item We introduce a lightweight Temporal Context Mask Refinement module to replace the conventional VAE feature extractor. This module effectively optimizes segmentation mask quality with low computational overhead.
\item  Our model achieves state-of-the-art performance on four public RVOS benchmarks.
\end{itemize}
\section{Related work}
Unlike Semi-supervised Video Object Segmentation (SVOS)~\cite{XMem, DEVA, cutie}, which relies on the ground truth annotation of the first frame to guide the segmentation of subsequent frames, Referring Video Object Segmentation (RVOS) utilizes natural language descriptions to localize and segment the referred target objects across the entire video sequence. This paradigm shift imposes higher demands on the model's ability to comprehend textual semantics and establish robust cross-modal alignments between video and language representations.
Early works in RVOS, such as MTTR~\cite{MTTR} and ReferFormer~\cite{ReferFormer}, typically adopted the Swin Transformer as the backbone for visual feature extraction and incorporated CondInst for instance segmentation due to its lightweight structure and adaptability. Building upon this architecture, SgMg~\cite{SgMg} introduced a Spectrum-guided Cross-modal Fusion mechanism to enhance the alignment between textual and visual modalities. It also appended a mask refinement module to alleviate the issue of feature drift, which commonly occurs in long video sequences.
VD-IT~\cite{VD-IT}, innovatively brought pre-trained text-to-video diffusion models into the RVOS pipeline. By leveraging the powerful U-Net structure of diffusion models, VD-IT achieved impressive performance in modeling spatiotemporal features, highlighting the untapped potential of generative architectures in this domain.\\
ReferDINO~\cite{ReferDINO} was built upon the GroundingDINO~\cite{GroundingDNIO} framework and further introduced a temporal enhancement module to improve tracking stability and temporal coherence across frames. Moreover, ReferDINO compares existing segmentation approaches and introduces some improvements, though the overall changes remain relatively limited. SAMWISE~\cite{samwise} builds upon SAM2~\cite{SAM2} by incorporating lightweight multimodal adapters and explicit temporal modeling. Without requiring any fine-tuning of the original model weights, it significantly enhances the system’s language understanding and temporal awareness, enabling high-performance object segmentation in streaming scenarios.
Although recent advances have improved multimodal fusion, visual-textual representation learning, and temporal reasoning, the core component of instance segmentation has seen limited innovation. Most RVOS frameworks still rely on conventional segmentation strategies with minimal refinement, despite their crucial impact on final mask quality.
\section{Method}
\begin{figure*}
\centering
\includegraphics[width=0.95\textwidth]{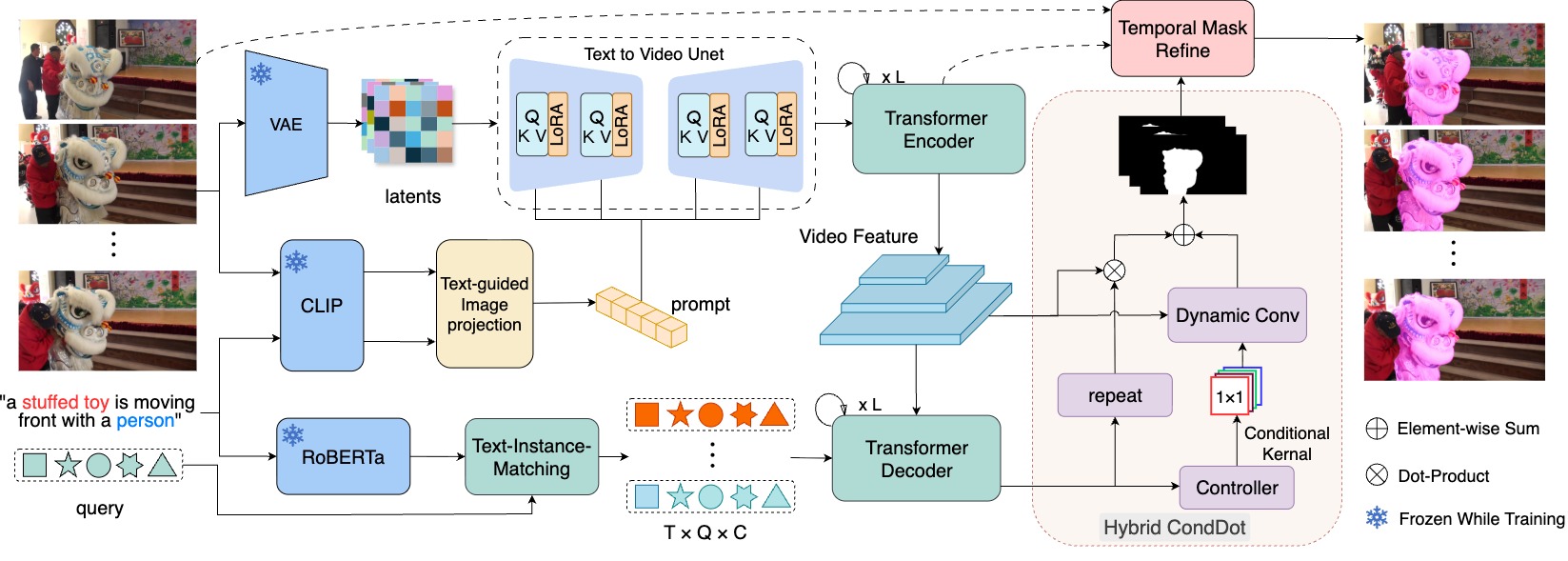} 
\caption{The model is built upon the DETR architecture. Its core components include: a feature extractor, a Transformer encoder, a Transformer decoder, and an HCD module. Specifically, the feature extractor and Transformer encoder work cooperatively to extract video features conditioned on the textual description. The Transformer decoder is then responsible for decoding target information utilizing the textual description. Finally, HCD generates the segmentation mask by integrating the decoded target information with the video features.}
\label{fig2}
\end{figure*}
\subsection{Overall Pipeline}
We adopt VD-IT~\cite{VD-IT} as our baseline framework, as the methods it introduces still hold considerable untapped potential for further exploration and optimization. As shown in Figure 2, the overall architecture of our proposed model is structured as follows: Given an input video clip consisting of $T$ frames, along with a corresponding textual description serving as a conditional prompt, the model is designed to generate pixel-wise segmentation masks for the entire sequence. Specifically, the output is a set of masks $M \in \mathbb{R}^{T \times H \times W}$, where $H$ and $W$ represent the height and width of each video frame. The goal is to accurately segment the referred object throughout the video, conditioned on the semantic cues provided by the text input.
\subsection{Noise-Free Text-to-Video Diffusion Model}
For feature extraction, we replace the Video Swin Transformer \cite{SwinTransformer,videoSwinTransformer} with pre-trained text-to-video diffusion models~\cite{videocrafter2,modelscope}. This choice is driven by recent VD-IT results showing that diffusion models offer superior temporal consistency, which is critical for video understanding tasks such as RVOS. Moreover, compared to Video Swin Transformer, diffusion models demonstrate markedly greater robustness in challenging dynamic scenarios, including drastic lighting changes, occlusions, and complex backgrounds.\\
Specifically, the input video frames are first encoded by VAE, which produces a compact visual representation organized into $F_l=\{f_{l,t}\}_{t=1}^T, f_l \in \mathbb{R}^{\frac{H}{8} \times \frac{W}{8} \times 4}$, where $H$ and $W$ represent the height and width of the input frames, respectively. In parallel, the accompanying text description is processed by the CLIP \cite{clip} text encoder, which transforms the textual input into a high-dimensional semantic embedding. To achieve effective cross-modal understanding, visual features extracted from the same video frames by the CLIP image encoder attend to the textual embeddings through a cross-modal attention mechanism. This operation aligns the visual and linguistic modalities more effectively, producing a refined, modality-aware prompt that encodes both the visual context and semantic intent of the description.
The fused prompt, together with the latent visual features $F_l$, is then fed into a diffusion model. Unlike conventional pipelines that treat spatial reasoning and temporal reasoning as separate modules, the diffusion model conducts joint spatiotemporal reasoning, enabling seamless fusion across time while preserving spatial details. This unified modeling approach allows the system to effectively capture complex, temporally-evolving dynamics across frames, resulting in a temporally coherent and semantically enriched video feature representation. \\
In conventional pipelines, the latent feature $F_l$ is weighted and combined with a Gaussian noise tensor before entering the diffusion model. However, our empirical observations reveal that the injected noise introduces instability into the segmentation results. Since segmentation tasks are highly sensitive to the precision of input features, we experimentally remove the noise prediction module. Surprisingly, this modification not only improves the stability of the outputs but also maintains the original segmentation accuracy, all while simplifying the overall architecture of the model.
\subsection{Query Instances}
First, we randomly initialize $Q$ learnable query instance vectors $q \in \mathbb{R}^{Q \times C}$ as the initial query representations, where $C$ denotes the channel dimension. Then, RoBERTa~\cite{RoBERTa} is employed to extract word-level features $F_w \in \mathbb{R}^{L \times C}$ from the input text, where $L$ denotes the length of features. Through a cross-attention mechanism, the query vectors $q$ are fused with the word features $F_w$ to generate semantically enhanced query vectors $q_w \in \mathbb{R}^{Q\times C}$. Fianlly, $q_w$ and the video feature map are fed into the Transformer decoder to obtain $Q \in \mathbb{R}^{Q \times C}$, which serve as semantic information for subsequent segmentation. In both the cross-attention and Transformer decoder stages, the query instances serve as queries, whereas the remaining elements serve as keys and values. In addition to mask prediction, $Q$ is also passed through linear layers to predict bounding boxes and classifications. The classification indicates whether the object of interest is present in the current frame, while the bounding box specifies the object’s location.
\subsection{Hybird CondDot}
\begin{figure*}[t]
\centering
\includegraphics[width=0.95\textwidth]{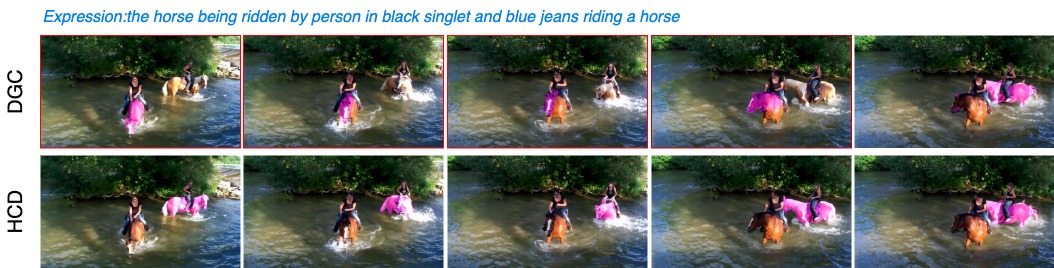} 
\caption{"HCD" denotes the Hybrid CondDot-based segmentation method. "DGC denotes the Dot Guided CondInst-based segmentation method."}
\label{fig4}
\end{figure*}
\begin{figure}[t]
\centering
\includegraphics[width=0.9\columnwidth]{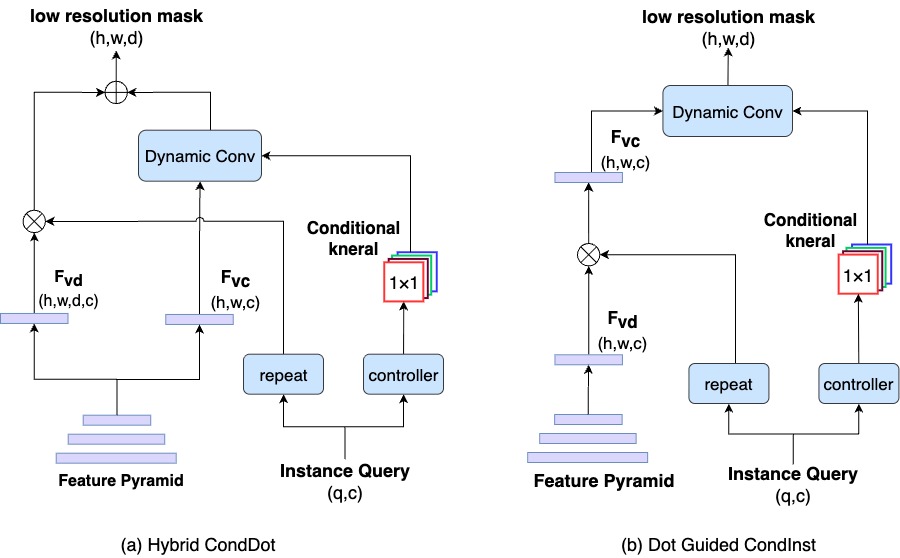} 
\caption{Detail illustrate of Hybrid CondDot and Dot Guided CondInst}
\label{fig3}
\end{figure}
As noted above, neither dot-product mechanisms nor CondInst frameworks are fully capable of capturing the complex semantic structures required for accurate segmentation. To better capture high-level semantics while retaining fine-grained details, we explored fusion strategies integrating these two methods. As illustrated in Figure~\ref{fig3}, following this design philosophy, we propose two complementary model architectures: Hybrid CondDot (HCD) and Dot Guided CondInst (DGC).
HCD adopts a parallel fusion strategy. In this architecture, the dot-product branch and the CondInst branch independently generate low-resolution mask predictions. The outputs from these two branches are subsequently fused via element-wise summation, enabling the model to effectively integrate complementary features extracted by the two methods. This preserves the global semantic perception provided by the CondInst while maintaining the boundary refinement capability of dot-product.\\
DGC employs a cascaded fusion strategy. This approach first leverages the dot-product to globally enhance low-level features, extracting information-rich and semantically potent representations. These enhanced features are then fed into the CondInst conditional instance segmentation module, which performs progressive refinement to predict instance-level masks. In this way, the global semantic information supplied by the dot-product mechanism can effectively guide the CondInst framework to focus on semantically critical regions during the prediction process, thereby improving both segmentation accuracy and stability.\\
More specifically, we get instance queries $Q$ from Transformer Decoder. Then we get 3-level video features downsampled by factors of $8 \times$, $16 \times$ and $32 \times$ from Transformer Encoder. These instance queries and multi-scale video features serve as the inputs to the segmentation head. $F_v \in \mathbb{R}^{\frac{H}{8}\times\frac{W}{8}\times C}$ denotes the result of resampling all video features to the spatial resolution corresponding to an 8× downsampling, followed by summation. The formula for HDC is
\begin{equation}
   M_n=F_{v}W^{D}repeat(Q)+Conv(F_{v}W^{C}, QW^{Q}),
\end{equation}
the formula for DGC is 
\begin{equation}
M_n=Conv(F_{v}repeat(Q)W^{C}, QW^{Q}),
\end{equation}
where $W^{D}$, $W^{C}\in \mathbb{R}^{C \times d^2}$ and $W^{Q}\in\mathbb{R}^{C \times N_K}$ are learnable parameters, $d$ is the upsampling rate and $N_k$ denotes denotes the sum of the numbers of convolution kernels and biases. \\
As illustrated in Figure~\ref{fig4}, the two approaches, DGC and HCD, exhibit notable differences in segmentation performance, particularly in terms of temporal coherence and spatial completeness. Specifically, DGC suffers from a significant temporal delay in target identification, correctly recognizing the target object only in the final frame of the video sequence. This delayed response suggests inadequate perception and semantic understanding during earlier frames. In addition, the segmentation masks produced by DGC in preceding frames consistently show substantial omissions in the target regions, indicating a limited ability to leverage decoded information for effective spatial modeling.
In comparison, HCD demonstrates greater robustness and accuracy in both recognition and segmentation. It consistently identifies the target object across all frames without exhibiting temporal delays, reflecting stronger temporal reasoning capabilities. More importantly, the segmentation masks generated by HCD display precise boundaries and complete target coverage, substantially outperforming those produced by DGC in terms of spatial quality and consistency.
Based on this comparative analysis, we choose HCD as our final segmentation method due to its superior performance in temporal consistency, segmentation precision, and overall robustness.\\
After obtaining $M_n \in \mathbb{R}^{Q \times \frac{H}{8} \times \frac{W}{8} \times d^2}$, where $Q$ denotes the number of queries, we further employ the Hungarian algorithm during training to perform optimal matching, selecting from all instance queries the ones that best correspond to the ground truth targets, thereby obtaining the precisely aligned low-resolution mask $M_l \in \mathbb{R}^{\frac{H}{8} \times \frac{W}{8} \times d^{2}}$.

\subsection{Temporal Context Mask Refinement}
\begin{table*}[t]
\centering
\begin{tabular}{l c c c c c c c c}
\toprule
\multirow{2}{*}{\textbf{Method}} & \multirow{2}{*}{\textbf{Publication}} & \multirow{2}{*}{\textbf{FPS}} & \multicolumn{3}{c}{\textbf{Ref-YouTube-VOS}} & \multicolumn{3}{c}{\textbf{Ref-DAVIS17}}
\\ \cmidrule(lr){4-6} \cmidrule(lr){7-9}
& & & $\mathcal{J\&F}$ & $\mathcal{J}$ & $\mathcal{F}$ & $\mathcal{J\&F}$ & $\mathcal{J}$ & $\mathcal{F}$ \\
\midrule
CMSA \cite{CMSA}               & CVPR'19 & - & 36.4 & 34.8 & 38.1 & 40.2 & 36.9 & 43.5 \\
URVOS \cite{URVOS}             & ECCV'20 & - & 47.2 & 45.3 & 49.2 & 51.5 & 47.3 & 56.0 \\
ReferFormer \cite{ReferFormer} & CVPR'22 & 50 & 62.9 & 61.3 & 64.6 & 61.1 & 58.1 & 64.1 \\
MLRL \cite{MLRL}               & CVPR'22 & - & 49.7 & 48.4 & 51.0 & 52.8 & 50.0 & 55.4 \\
SgMg \cite{SgMg}               & ICCV'23 & 65 & 65.7 & 63.9 & 67.4 & 63.3 & 60.6 & 66.0 \\
VD-IT \cite{VD-IT}             & ECCV'24 & 21 & 66.5 & 64.4 & 68.5 & 69.4 & 66.2 & 72.6 \\
LoSh \cite{LoSh}               & CVPR'24 & - & 67.2 & 65.4 & 69.0 & 64.3 & 61.8 & 66.8 \\
DsHmp \cite{DsHmp}             & CVPR'24 & - & 67.1 & 65.0 & 69.1 & 64.9 & 61.7 & 68.1 \\
SAMWISE \cite{samwise}         & CVPR'25 & - & 69.2 & 67.8 & 70.6 & 70.6 & 67.4 & 74.5 \\
SSA \cite{SSA}                    & CVPR'25 & - & 64.3 & 62.2 & 66.4 & 67.3 & 64.0 & 70.7 \\
DMVS \cite{DMVS}                   & CVPR'25 & - & 64.3 & 62.4 & 66.2 & 65.2 & 62.2 & 68.2 \\
ReferDINO \cite{ReferDINO}     & ICCV'25 & - & 69.3 & 67.0 & 71.5 & 68.9 & 65.1 & 72.9 \\
\midrule
\textbf{HCD (Ours)} & This work & 19 &\textbf{71.1} & \textbf{68.9} & \textbf{73.3} & \textbf{71.2} & \textbf{67.5} & \textbf{75.0} \\
\bottomrule
\end{tabular}
\caption{Comparison with state-of-the-art methods on the validation sets of Ref-YouTube-VOS and Ref-DAVIS17 datasets.}
\label{tab:main_result}
\end{table*}
\begin{table*}[t]
\centering
\begin{tabular}{l c c c c c c c c}
\toprule
\multirow{2}{*}{\textbf{Method}} & \multirow{2}{*}{\textbf{Publication}} & \multicolumn{3}{c}{\textbf{A2D-Sentences}} & \multicolumn{3}{c}{\textbf{JHMDB-Sentences}}
\\ \cmidrule(lr){3-5} \cmidrule(lr){6-8}
& & $\textbf{mAP}$ & $\textbf{oIoU}$ & $\textbf{mIoU}$ & $\textbf{mAP}$ & $\textbf{oIoU}$ & $\textbf{mIoU}$ \\
\midrule
ReferFormer \cite{ReferFormer} & CVPR'22 & 52.8 & 77.6 & 69.6 & 42.2 & 71.9 & 71.0 \\
SgMg \cite{SgMg}               & ICCV'23 & 56.1 & 78.0 & 70.4 & 44.4 & 72.8 & 71.7 \\
VD-IT \cite{VD-IT}             & ECCV'24 & 61.4 & 81.5 & 73.2 & 46.5 & 74.4 & 73.4 \\
LoSh \cite{LoSh}               & CVPR'24 & 59.9 & 81.2 & 73.1 & - & - & - \\
DsHmp \cite{DsHmp}             & CVPR'24 & 59.8 & 81.1 & 72.9 & 45.8 & 73.9 & 73.0 \\
ReferDINO \cite{ReferDINO}     & ICCV'25 & 61.1 & \textbf{82.1} & 73.6 & 46.6 & 74.2 & 73.2 \\
\midrule
\textbf{HCD (Ours)} & This work & \textbf{62.5} & 82.0 & \textbf{74.2} & \textbf{47.1} & \textbf{75.2} & \textbf{73.9} \\
\bottomrule
\end{tabular}
\caption{Comparison with state-of-the-art methods on the validation sets of A2D-Sentences, JHMDB-Sentences datasets.}
\label{tab:a2d_result}
\end{table*}
Upon obtaining the low-resolution mask $M_l$, our method performs mask refinement and spatial resolution enhancement by fusing $8 \times$ and $4 \times$ downsampled video features. 
Specifically, $M_l$ is first concatenated with multi-scale video features along the channel dimension, followed by feature compression and information integration via low-dimensional projection. 
The restored mask is then upsampled from $\mathbb{R}^{\frac{H}{8} \times \frac{W}{8} \times d^2}$ to $\mathbb{R}^{\frac{Hd}{8} \times \frac{Wd}{8}}$ to generate the high-resolution mask. $F_{\times 8}$ denots $8 \times$ downsampled features directly extracted from the diffusion model and $F_{\times 4}$ denotes 4× downsampled features. Rather than generating $F_{\times 4}$ through the VAE, which has limited capacity capture temporal dependencies in video sequences, we obtain $F_{\times 4}$ by applying a 4× downsampling operation directly to the raw video frames and employ a temporal convolution module to explicitly model inter-frame motion. This approach leads to notable improvements in temporal coherence during mask reconstruction. The formula is
\begin{equation}
    M=Up_{\times 4}(cnov(up_{\times 2}(conv(M_{l}\oplus F_{\times 8})) \oplus F_{\times 4})),
\end{equation}
where $\oplus$ denotes concatenation along the channel dimension, $up_2$ and $up_4$ represent upsampling by factors of 2 and 4, respectively.
\section{Experiment}
\subsection{Datasets and Evaluation Metrics}
We conduct comprehensive evaluations of the proposed HCD model on four widely used benchmarks: Ref-YouTube-VOS~\cite{youtube}, Ref-DAVIS17~\cite{davis}, JHMDB-Sentences~\cite{a2d}, and A2D-Sentences~\cite{a2d}. These datasets collectively cover a diverse range of challenges in referring video object segmentation, including variations in motion, appearance, and linguistic expression. To ensure a fair and standardized comparison, we adopt commonly used evaluation metrics: region similarity (J), which measures the average Intersection-over-Union (IoU) between predicted and ground truth masks; contour accuracy (F), which assesses boundary-level agreement; and their average (J\&F), which serves as an overall indicator of segmentation quality. \\
\subsection{Implementation Details}
Our model leverages the pre-trained text-to-video diffusion model ModelScopeT2V~\cite{modelscope} as the visual backbone and RoBERTa as the textual backbone. The backbones of ModelScopeT2V are kept frozen during training, while the Transformer is fine-tuned using Low-Rank Adaptation (LoRA) with a rank of 16. Both the Transformer encoder and decoder adopt the Deformable Transformer~\cite{deformable} architecture.
We initially train our model on Ref-YouTube-VOS, followed by fine-tuning on Ref-DAVIS17, A2D-Sentences, and JHMDB-Sentences before evaluating on the corresponding test sets. The number of instance queries $Q$ is set to 5. Training is performed on two NVIDIA A100 GPUs (40GB each), taking approximately four days to complete eight epochs.
While previous works typically report results after pre-training on Ref-COCO, we omit this step. Our model, when trained solely on Ref-YouTube-VOS, already achieves strong and competitive performance, demonstrating the effectiveness of our approach without the need for additional pre-training. \\
\textbf{Loss Function.} We compute the losses for mask predictions, confidence scores, and bounding boxes individually, and derive the final loss by taking a weighted summation of these components. Specifically, Dice loss~\cite{dice} and Focal loss~\cite{focalLoss} are applied to the mask predictions, Focal loss is used for the confidence scores, and a combination of L1 loss and GIoU loss~\cite{GIoU} is employed for bounding boxes.
\subsection{Comparison with SOTA Method}
We conduct a comprehensive comparison between our proposed HCD framework and several state-of-the-art approaches, as summarized in Table~\ref{tab:main_result}. Across both Ref-YouTube-VOS and Ref-DAVIS17 benchmarks, HCD consistently surpasses existing methods by a clear margin.
Specifically, on Ref-YouTube-VOS, our framework achieves a $J\&F$ score of 71.1, outperforming the previous best result by 2.4 points. Notably, even without pre-training on the Ref-COCO dataset, HCD still surpasses VD-IT by 4.6 points, underscoring the inherent capability of our model to deliver competitive performance without relying on extensive external supervision. These findings highlight the effectiveness of our approach, particularly in improving segmentation quality with respect to boundary precision and instance-level temporal consistency.
On Ref-DAVIS17, HCD obtains a $J\&F$ score of 71.2, exceeding the previous state-of-the-art by 0.6 points. Although the margin is relatively smaller, it further validates the robustness and strong generalization ability of our method across different datasets and diverse video scenarios. These results collectively demonstrate that the proposed HCD framework offers a substantial and consistent advancement in referring video object segmentation.\\
We also evaluate the HCD method on the A2D-Sentences and JHMDB-Sentences datasets. As shown in Table~\ref{tab:a2d_result}, HCD surpasses the current state-of-the-art methods by 1.1 and 0.8 point on A2D-Sentences and JHMDB-Sentences, respectively. These results not only confirm the superiority of HCD in conventional tasks, but also demonstrate its strong generalization capability and robust performance in action-related semantic understanding and dynamic scene modeling.
\subsection{Ablation Studies}
\begin{table}[t]
    \centering
    \begin{tabular}{c c c c}
    \toprule
         \multirow{2}{*}{\textbf{Segmentation Head}} & \multicolumn{3}{c}{\textbf{Performance}} \\
         \cmidrule(lr){2-4}
         & $\mathcal{J\&F}$ & $\mathcal{J}$ & $\mathcal{F}$ \\
         \midrule
         Dot product & 69.7 & 67.5 & 71.9 \\
         CondInst & 70.4 & 68.3 & 72.6 \\
         DGC & 69.7 & 67.6 & 71.8 \\
         HCD & \textbf{71.1} & \textbf{68.9} & \textbf{73.3} \\
         \bottomrule
    \end{tabular}
    \caption{Ablation study of Segmentation Head method.}
    \label{tab:ablation_seg}
\end{table}
\begin{table}[t]
    \centering
    \begin{tabular}{c c c c c}
    \toprule
          \multirow{2}{*}{\textbf{Mask Refinement}} & \multirow{2}{*}{\textbf{NP}} & \multicolumn{3}{c}{\textbf{Performance}} \\
         \cmidrule(lr){3-5}
         & & $\mathcal{J\&F}$ & $\mathcal{J}$ & $\mathcal{F}$ \\
         \midrule
         VAE          & \ding{51} & 61.1          & 58.3          & 63.9 \\
         VAE          & \ding{55} & 61.3          & 58.5          & 64.0 \\
         CNN          & \ding{51} & 61.3          & 58.4          & 64.3 \\
         CNN          & \ding{55} & 61.5          & 58.6          & 64.4 \\
         TCMR          & \ding{51} & 61.9          & 59.1          & 64.8 \\
         TCMR          & \ding{55} & \textbf{62.2} & \textbf{59.3} & \textbf{65.1} \\
         \bottomrule
    \end{tabular}
    \caption{Ablation study of Mask Refinement method and Noise Prediction. "NP" denotes Noise Prediction.}
    \label{tab:ablation_refine}
\end{table}
We perform ablation experiments on Ref-Youtube-VOS and Ref-DAVIS17 to examine the effect of critical components in our model. Table \ref{tab:ablation_seg} present results on the Ref-YouTube-VOS, and Table \ref{tab:ablation_refine} on the DAVIS17.
\subsubsection{Segmentation Head Analysis.} 
As shown in Table~\ref{tab:ablation_seg}, we try four different methods for the segmentation head. Dot-product directly multiplies each query embedding with the feature map to obtain the mask, while CondInst utilizes each query to generate conditional kernels and corresponding biases, then performs convolution on the feature map. HCD and DGC are the two new components we previously proposed. The experimental results show that although DGC is designed to integrate the dot-product mechanism with CondInst, its performance falls short of CondInst and is only comparable to the dot-product method. This phenomenon suggests that when adopting a serial fusion strategy, information transmitted between different modules may be weakened or lost to some extent, resulting in critical information not being fully preserved or effectively utilized. Due to interruptions or dilution in the information flow, the model struggles to fully leverage the strengths of each component, ultimately leading to degraded overall performance and instability. In contrast, HCD significantly improves segmentation performance through a parallel fusion strategy, achieving an increase of 0.7 point over CondInst and 1.4 point over the dot-product method on $J\&F$.\\
These results fully demonstrate the effectiveness of the HCD design. The parallel architecture of HCD not only effectively mitigates information loss during transmission but also more comprehensively integrates the advantageous features of both methods while maintaining a simple and efficient model structure. Consequently, HCD demonstrates clear advantages in enhancing segmentation accuracy and boundary precision, thereby substantiating its superiority and potential as a segmentation head design.
\subsubsection{Mask Refinement and Noise Prediction Analysis.}
We systematically evaluate the effectiveness of the proposed TCMR through comparative experiments with CNN and VAE. As shown in Table~\ref{tab:ablation_refine}, replacing the VAE with a standard CNN yields a modest performance improvement, with the $J\&F$ increasing by 0.2 percentage point. This indicates that the VAE’s feature extraction capability is relatively limited for this task and struggles to fully capture the target features. Subsequently, replacing the CNN with TCMR results in a significant performance gain, with the $J\&F$ improving by 0.6 percentage point. This further validates the importance of temporal modeling capability over mere feature extraction in mask refinement tasks. In summary, the TCMR module effectively integrates temporal contextual information, substantially enhancing mask refinement performance and providing a more accurate and robust solution for video object segmentation. Additionally, we removed the noise prediction module from all methods and observed consistent improvements in performance, indicating that this component is redundant within the current architecture and can adversely affect model accuracy.
\subsection{Temporal Consistency Analysis.}
\begin{figure}[t]
\centering
\includegraphics[width=0.9\columnwidth]{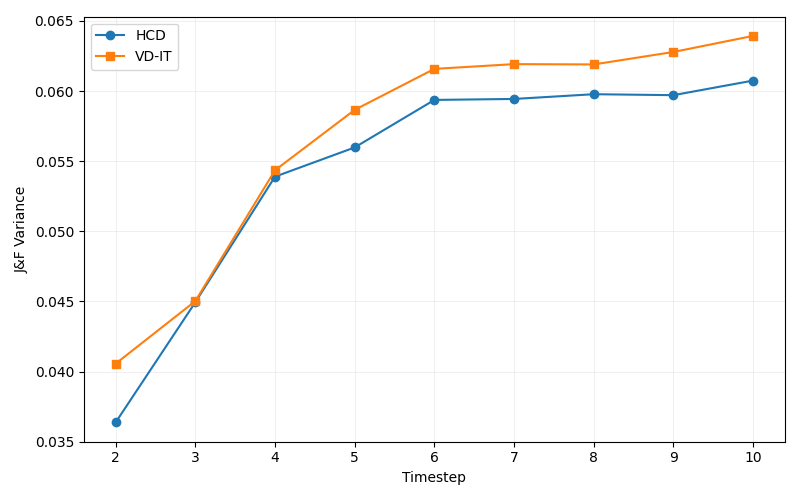} 
\caption{Temporal Consistency Analysis. All videos from the Ref-YouTube-VOS test set were sampled, and the variance of $J\&F$ between the initial frame and the following 10 frames was reported.}
\label{fig5}
\end{figure}
As shown in Figure~\ref{fig5}, we compare the $J\&F$ variance between 2 to 10 consecutive frames for VD-IT and HCD. The growth of $J\&F$ variance across consecutive video frames is slower and more stable for HCD compared to VD-IT. This indicates that HCD has a stronger advantage in temporal stability, maintaining more consistent segmentation performance throughout the video sequence. The slower increase in $J\&F$ variance clearly suggests that the model is more robust and resilient to inter-frame variations and less susceptible to challenging factors such as object deformation, occlusion, or background interference, thereby demonstrating HCD’s superior temporal consistency. This significant performance gain is primarily attributed to the architectural improvements in HCD. Its segmentation approach is carefully designed to more effectively exploit and integrate both spatiotemporal information and semantic information embedded in the feature maps. By introducing stronger and more comprehensive contextual modeling capabilities during the target representation extraction process, HCD effectively ensures higher temporal consistency and long-term stability of segmentation results across different timesteps. In contrast, VD-IT demonstrates relatively weaker alignment and robust fusion of features along the temporal dimension, which leads to larger fluctuations between consecutive frames and consequently results in higher variance and reduced segmentation reliability.
\subsection{Qualitative Analysis}
\begin{figure}[t]
\centering
\includegraphics[width=0.95\columnwidth]{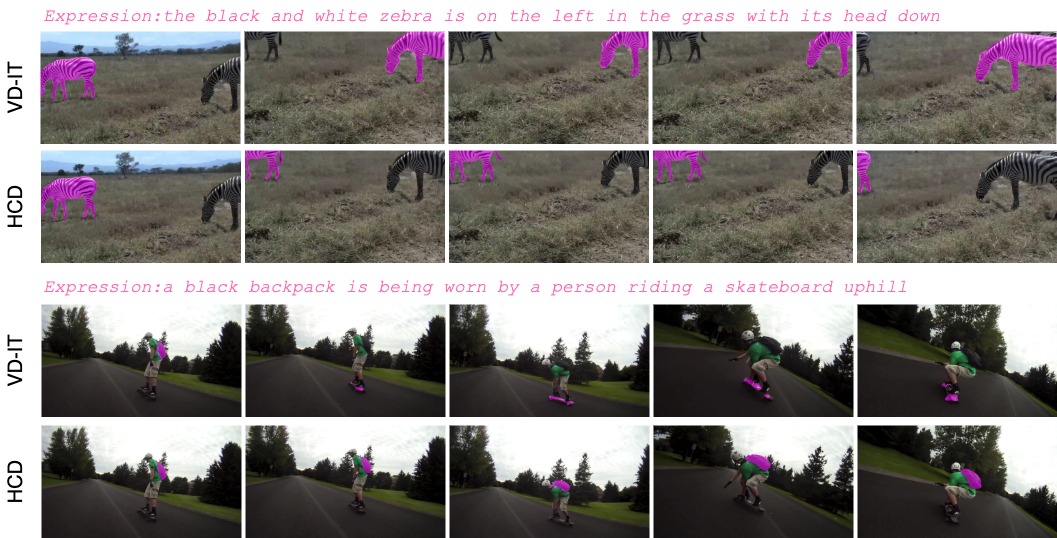} 
\caption{Qualitative comparison of HCD with VD-IT}
\label{fig6}
\end{figure}
\subsubsection{Qualitative Comparison.} We compare the qualitative results of our HCD with VD-IT in Figure~\ref{fig6}. VD-IT is able to achieve accurate target identification and segmentation in the initial frame and provides a precise starting point for subsequent predictions. However, as the temporal sequence progresses, particularly when encountering complex scenarios such as target occlusion or temporary loss, this method exhibits significant performance degradation and is prone to object confusion or misidentification. In contrast, HCD demonstrates superior temporal robustness. The key lies in its effective modeling and use of both spatio-temporal consistency and contextual information across frames. This capability enables HCD to stably maintain precise segmentation of the target region even under extreme conditions involving severe occlusion, thereby dramatically enhancing the continuity and reliability of the segmentation results.
\section{Conclusion}
In this paper, we proposed HCD, a novel segmentation method for RVOS that integrates existing segmentation methods, leverages rich feature information to predict more stable and accurate masks. Furthermore, based on our experimental analysis, we removed the noise prediction module within the model architecture, which improves the stability of the output features and simplifies the model structure. We also proposed a dedicated temporal context mask refinement module to mitigate the inherent insufficiency of temporal feature modeling during the mask refinement process. Finally, across all four major RVOS benchmarks, our approach consistently achieves state-of-the-art performance.

\bibliography{aaai2026}
\clearpage
\appendix
\section{Appendix}
\subsection{Experimental Details}
Regarding the model configuration, both the Transformer encoder and decoder are composed of 4 layers, with the number of queries fixed at 5. During training, the Hungarian algorithm is utilized to establish an optimal matching between predicted queries and ground truth targets, thereby assigning each target instance to the most appropriate query. This approach not only facilitates precise loss computation but also effectively mitigates matching ambiguities, promoting more efficient and stable model convergence. In the inference stage, the Hungarian algorithm is omitted, and the outputs of all queries are aggregated via element-wise summation. This aggregation strategy harnesses the complementary information captured by multiple queries, effectively reducing the instability associated with individual predictions, and yields a smoother, more robust final segmentation result.


\subsection{Ablation Results on the number of Encoder and Decoder Layers}
we conduct the ablation study on Ref-YouTube-VOS. The Transformer Encoder is responsible for extracting rich and multi-level spatiotemporal features from the input video, accurately capturing the dynamic changes of target objects and contextual information. This enables the model to gain a deep understanding of the temporal relationships within the video content, thereby laying a solid foundation for subsequent segmentation tasks. Meanwhile, the Transformer Decoder generates query embeddings based on the input textual description. These embeddings serve as conditional signals to guide the segmentation process, allowing the model to precisely focus on the target regions specified by the text. As shown in Table~\ref{tab:ablation_layer}, reducing the number of encoder layers leads to a significant performance drop, whereas increasing the number of layers yields negligible improvement. This indicates that setting the encoder depth to four layers achieves an optimal balance, as fewer layers noticeably weaken the feature extraction capability, while additional layers fail to capture more informative features.\\
Subsequently, we experimented with both increasing and decreasing the number of decoder layers, and observed that the model performance exhibited only a slight decline in both cases. This suggests that the performance bottleneck no longer lies in the quality of the query embeddings, but rather in how effectively the segmentation head leverages these embeddings. In other words, further performance improvement depends more on optimizing the segmentation head’s ability to interpret and utilize the query embeddings, rather than simply adjusting the number of decoder layers.
\begin{table}
    \centering
    \begin{tabular}{c c c c c}
    \toprule
         \textbf{Decoder Layer} & \textbf{Encoder Layer} & $\mathcal{J\&F}$ & $\mathcal{J}$ & $\mathcal{F}$ \\
         \midrule
            4 & 4 & 70.6 & 68.5 & 72.8 \\
            5 & 4 & 69.7 & 67.5 & 71.9 \\
            3 & 4 & 69.1 & 66.8 & 71.4 \\
            4 & 3 & 67.7 & 65.7 & 69.8 \\
            4 & 5 & 70.6 & 68.5 & 72.7 \\
         \bottomrule
    \end{tabular}
    \caption{Ablation study of Decoder and Encoder layers.}
    \label{tab:ablation_layer}
\end{table}

\subsection{Ablation Results on Timestep}
\begin{table}[t]
    \centering
    \begin{tabular}{c c c c c}
    \toprule
         \textbf{Timestep} & \textbf{weight} & \textbf{Learnable Weight} & $\mathcal{J\&F}$ \\
         \midrule
         0          & 2.92 \%     & \ding{55} & 54.89  \\
         1          & 4.13 \%     & \ding{55} & 54.61  \\
         3          & 5.85 \%     & \ding{55} & 53.88  \\
         5          & 7.10 \%     & \ding{55} & 54.51  \\
         10         & 9.70 \%     & \ding{55} & 52.52  \\
         25         & 15.00\%    & \ding{55} & 52.94  \\
         50         & 22.00\%    & \ding{55} & 52.31  \\
         -          & 2.92 \%     & \ding{51} & 54.52  \\
         \bottomrule
    \end{tabular}
    \caption{Ablation study of timestep. weight denotes Noise Prediction. "weight" denotes the Gaussian noise weight corresponding to the respective timestep.}
    \label{tab:ablation_timestep}
\end{table}
\begin{figure*}[t]
\centering
\includegraphics[width=0.95\textwidth]{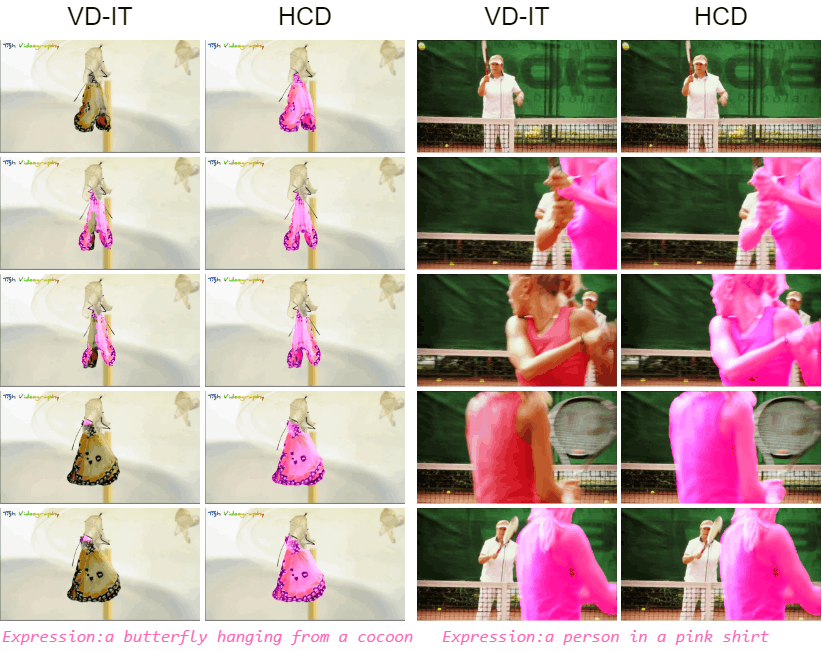} 
\caption{Qualitative comparison of our model with VD-IT}
\label{fig:qualitative}
\end{figure*}
When employing a diffusion model for video feature extraction, Gaussian noise corresponding to the selected timestep is inevitably introduced, and larger timesteps are associated with higher noise weights. We conducted an ablation study on the Ref-DAVIS17 dataset to investigate the impact of varying timesteps. As presented in Table~\ref{tab:ablation_timestep}, the overall model performance gradually decreases as the number of timesteps, corresponding to the noise weight, increases. Furthermore, we observe that the variance of the model’s output results also grows with more timesteps, indicating a progressively less stable performance. This phenomenon primarily arises from the interference of noise with the original feature representations. The introduction of noise compromises the integrity of these features, thereby diminishing the model’s capacity for precise target segmentation and resulting in a significant decline in segmentation accuracy.\\
Finally, we initialized the noise weight to its minimal value and treated it as a learnable parameter. However, the model’s performance remained inferior to the best results achieved with fixed weights, indicating that the inherent stochasticity of the noise significantly impacts the stable convergence of this parameter. Specifically, the random fluctuations introduced by the noise increase the uncertainty during training, making it challenging for the model to achieve effective and consistent optimization of this parameter, thereby limiting the potential for overall performance improvement. Therefore, for segmentation tasks, it is generally advisable to maintain a lower noise level, as this preserves the fidelity of feature representations, minimizes the detrimental impact of stochastic perturbations, and thereby enhances both the stability and accuracy of the segmentation outcomes.
\subsection{Additional Visualization Results}
To more comprehensively validate the performance advantages of our HCD, we present a visual comparison between HCD and the representative baseline VD-IT in Figure~\ref{fig:qualitative}. As illustrated, HCD consistently demonstrates superior performance in both semantic understanding and temporal consistency. In terms of semantics, HCD can more precisely identify and interpret the referred target, not only accurately delineating its overall contour but also capturing fine-grained local structures and boundary details. Even when the target exhibits complex shapes or highly irregular edges, HCD is able to maintain stable and accurate segmentation performance, thereby enhancing overall segmentation accuracy in various challenging scenarios. In terms of temporal consistency, the masks generated by HCD maintain high stability not only between adjacent frames but also demonstrate strong continuity across frames with larger temporal gaps, effectively reducing target loss caused by occlusions.

\end{document}